\documentclass[10pt, conference, letterpaper]{IEEEtran}

\usepackage{amsmath,amsthm,amssymb}
\usepackage[compress]{cite}
\usepackage{tabto}
\usepackage{algorithm}
\usepackage[noend]{algpseudocode}
\usepackage{xspace}
\usepackage{paralist}
\usepackage{url}
\usepackage{graphicx}
\usepackage{tabularx,comment,makecell}
\usepackage{dsfont}
\usepackage[dvipsnames]{xcolor}
\usepackage{epigraph}
\usepackage{dblfloatfix}

\newcommand{\Fig}[1]{Fig.~\ref{fig:#1}}
\newcommand{\Prop}[1]{Property~\ref{prop:#1}}

\newcommand{\Sec}[1]{Sec.~\ref{sec:#1}}
\newcommand{\Tab}[1]{Tab.~\ref{tab:#1}}
\newcommand{\Eq}[1]{(\ref{eq:#1})}

\newcommand{\Dc}{\mathcal{D}}
\newcommand{\Mc}{\mathcal{M}}
\newcommand{\Nc}{\mathcal{N}}

\newcommand{\Xc}{\mathcal{X}}
\newcommand{\Ic}{\mathcal{I}}

\newtheorem{property}{Property}
\begin{document}

\title{Dependable Distributed Training \\ of Compressed Machine Learning Models}

\author{
\IEEEauthorblockN{Francesco~Malandrino\IEEEauthorrefmark{1}\IEEEauthorrefmark{2}, Giuseppe~Di~Giacomo\IEEEauthorrefmark{3},
Marco Levorato\IEEEauthorrefmark{4}, Carla~Fabiana~Chiasserini\IEEEauthorrefmark{3}\IEEEauthorrefmark{1}\IEEEauthorrefmark{2}
}
\IEEEauthorblockA{
\IEEEauthorrefmark{1}CNR-IEIIT, Italy --
\IEEEauthorrefmark{2}CNIT, Italy --
\IEEEauthorrefmark{3}Politecnico di Torino, Italy --
\IEEEauthorrefmark{4}UC Irvine, USA
}
}

\maketitle

\begin{abstract}
The existing work on the distributed training of machine learning (ML) models has consistently overlooked the {\em distribution} of the achieved learning quality, focusing instead on its average value. This leads to a poor {\em dependability} of the resulting ML models, whose performance may  be much worse than expected. We fill this gap by proposing DepL, a framework for {\em dependable learning orchestration}, able to make high-quality, efficient decisions on (i) the data to leverage for learning, (ii) the models to use and when to switch among them, and (iii) the clusters of  nodes, and the resources thereof, to exploit. For concreteness, we consider as possible available  models a full DNN and its compressed versions. 
Unlike previous studies, DepL guarantees that a target learning quality is reached {\em with a target probability}, while keeping the training cost at a minimum. We prove that DepL has constant competitive ratio and polynomial complexity, and show that it outperforms  the state-of-the-art by over 27\% and closely matches the optimum.
\end{abstract}

\section{Introduction}
\label{sec:intro}

Machine learning (ML) models, and deep neural networks (DNNs) in particular, are becoming more capable, but also harder and more costly to train. This issue has been tackled through two complementary approaches:   distributed training and model compression. The former exploits the resources (e.g., computational) and data available at different nodes,
thus better distributing the training burden. 
The latter includes a wealth of different techniques (e.g., model pruning and knowledge distillation) allowing for the transfer of  knowledge   from a model to a (usually, simpler) one.

Distributed training and model compression have been variously combined together in the literature, allowing the used resources to adapt to the chosen model~\cite{abdelmoniem2021resource}, selecting the best model given the available resources~\cite{wang2019adaptive,wu2021fast}, and even performing mutual adaptation between the two \cite{noi-infocom23}. However, an aspect that has been overlooked so far, which we aim to investigate in this work, is the {\em dependability} of DNN training. Indeed, ML is increasingly used for mission-critical (and even safety-related) applications, hence  traditional approaches focusing on the expected  learning quality may be inadequate to present needs.

In this work, we aim at providing the right network support to make  DNN training {\em dependable}, i.e., to guarantee that a target learning quality (e.g., loss or accuracy) is reached by a target time and {\em with a target probability} -- at the minimum cost. To this end, we make the key observation that learning dependability is affected by two main factors: (i) the used {\em models} (i.e., full or compressed versions), and (ii) the quality of {\em training} (including the data generated by data sources as well as the nodes and the resources thereof that are used). \emph{Our work is the first to account for all these contributions in a holistic manner, and to leverage them all to make high-quality, efficient training decisions.}
Importantly, although we focus on model compression as the source of alternative models, the problem we pose and the solution we envision  extend to arbitrary models.

\begin{figure}
    \centering
    \includegraphics[width=0.4\textwidth]{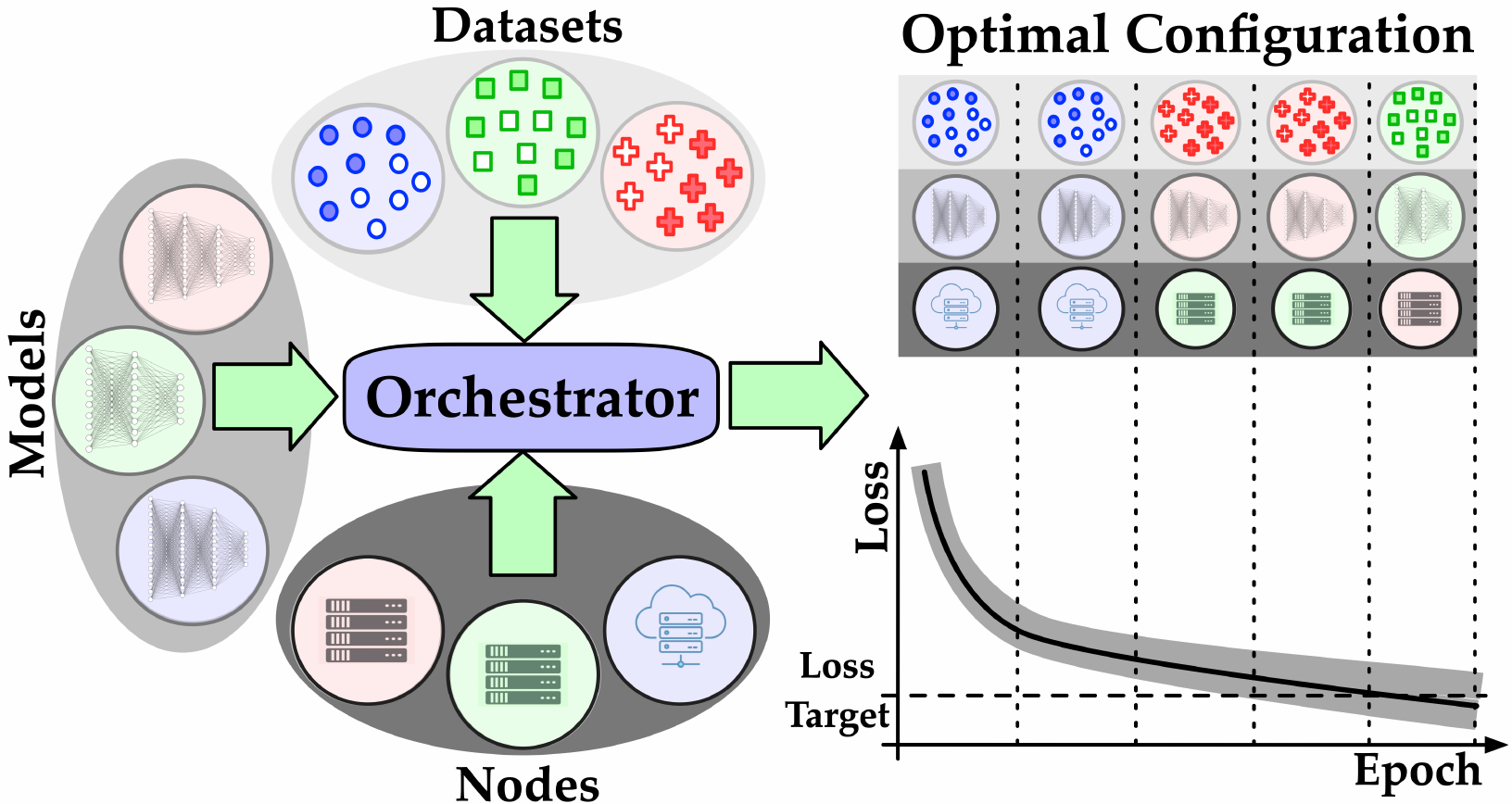}
\caption{
DepL enables the optimization of dependable, distributed training of neural networks over heterogeneous nodes, datasets, and model architectures. The optimal configuration is designed to control the uncertainty in the loss progression along training epochs.
}
\label{fig:plot_loss}
\vspace{-4mm}
\end{figure}
More specifically, we consider a distributed learning scenario where nodes in different parts of the network (e.g., far edge, edge, and cloud) cooperatively train a DNN model aided by a  learning orchestrator. In this setting, we develop and evaluate a novel DNN training scheme called DepL (for {\em dep}endable {\em l}earning), allowing the learning orchestrator to make joint decisions about: (a)
the data sources (hence, the datasets) to exploit for learning;
(b) the models to train for the ML task at hand, and when to switch from one to another;
(c) the node clusters to use at different stages of training. 
Unlike previous work, DepL accounts for the mutual interaction between such decisions, and the effect that each of them has on the expected learning quality {\em as well as its distribution}. Thanks to this ability, DepL can meet the challenges of present- and next-generation ML applications, including safety-critical ones. Also, DepL makes near-optimal decisions, minimizing the cost incurred by the learning process,  while keeping a remarkably low (polynomial)  computational complexity.
Finally, DepL can be combined with emerging approaches based upon {\em conformal prediction sets}~\cite{angelopoulos2022conformal,zecchin2023forking}, seeking to add reliability {\em a posteriori}, i.e., starting from an already-trained model. Indeed, models trained in a dependable manner through DepL also have smaller conformal sets.

The main contributions of our work are thus  as follows:

\textbullet~We propose a  comprehensive system model, capturing all relevant aspects of distributed ML training,   including learning quality distribution;

\textbullet~In light of the problem complexity, we propose a solution strategy called DepL, allowing to achieve the target learning quality  distribution at the minimum cost;

\textbullet~We analyze DepL's complexity and competitive ratio, proving the former to be polynomial and the latter to be both low and constant;

\textbullet~We show that DepL closely matches the optimum and consistently outperforms the state of the art. 

The rest of the paper is organized as follows. \Sec{motivation} further motivates our approach, considering a real-world example and showing why solely focusing on the expected learning quality may be  suboptimal.  \Sec{system-model} formally introduces our system model and problem formulation. The DepL solution  is then presented in \Sec{depl} and analyzed in \Sec{analysis}. Finally, after evaluating DepL's performance in \Sec{results} and discussing related work in \Sec{relwork}, we conclude the paper in \Sec{concl}.

\section{The Importance of  Dependable Training}\label{sec:motivation} 
We now illustrate the reasons why it is important to take a dependable approach to ML model training.   
As an example,  assume that the  popular AlexNet CNN model \cite{krizhevsky2017imagenet} for image classification has to be trained at the network edge, using the CIFAR-10 dataset \cite{krizhevsky2009learning}. To make it more amenable to the limited computational capability of edge servers, the model can be compressed by pruning it with a given factor. Clearly, the larger the pruning factor, the smaller and the less complex the model becomes, thus leading not only to a less demanding  model  for inference but also to a lower cost during training. On the other hand, a more compressed version of the model exhibits higher sensitivity to a drift in the distribution of the input data and may provide less accurate output during inference due to the reduced number of model weights \cite{blalock2020state,liebenwein2021lost}. 
{\em An aspect that however has been widely overlooked so far is how the dependability of training, i.e., the confidence level in the accuracy achieved  through training, may be severely impacted by the version of the model and the dataset that are used. Importantly, these, on their turn, depend upon the computational capability of the nodes executing the training as well as the data sources used for the training. } 
%
\begin{figure}
    \centering
    \includegraphics[width=0.35\textwidth]{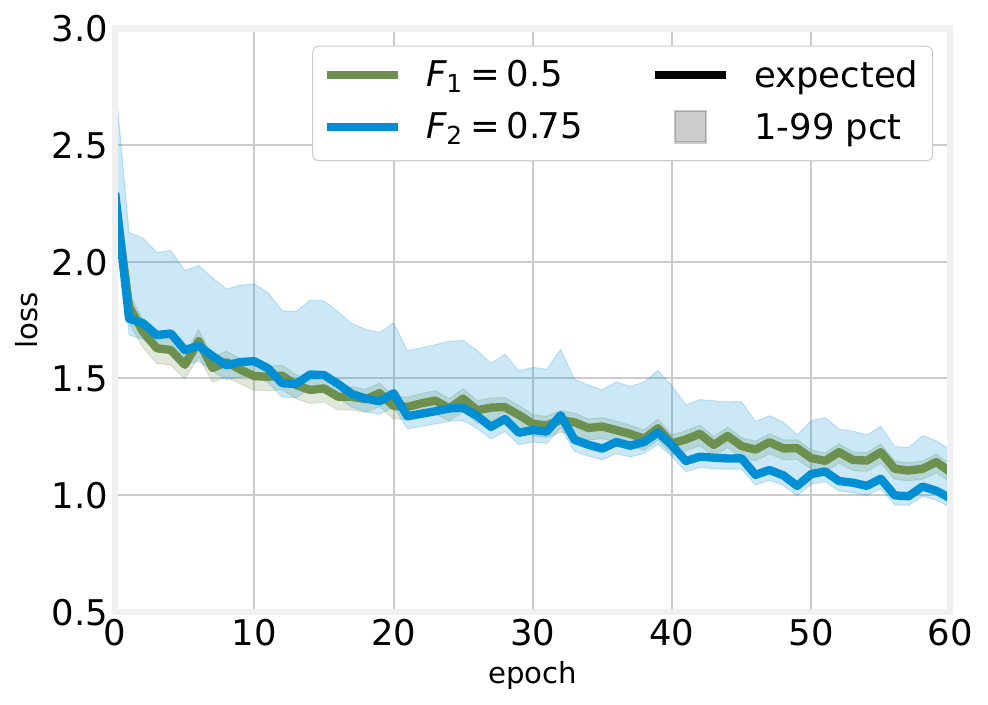}
\caption{
Example of evolution of the test loss during training of the AlexNet CNN model pruned with 0.5 and 0.75 factor. Lines represent the expected loss, while  shaded areas are delimited by the loss 1st and 99th percentiles.
}
\label{fig:plot_loss}
\vspace{-4mm}
\end{figure}

An example of experimental evidence of the above effect is given by the results  shown in  \Fig{plot_loss}, obtained using two different versions of the AlexNet model, one with pruning factor $F_1{=}0.5$ and the other with pruning factor $F_2{=}0.75$, and adopting the gradient descent optimizer with learning rate and  momentum, respectively, equal to $10^{-3}$ and $0.9$, and batch size equal to $64$.  
As a proxy metric for the achieved accuracy, the figure presents the evolution of the test loss through the training epochs. 
Three important facts can be noted:  
\begin{enumerate}
\item The trajectories of the test loss evolution over the epochs are not deterministic but stochastic in nature, as highlighted by the shaded areas; 
    \item Their behavior visibly depends on the model that is used;
    \item Additionally, the average and the empirical distribution of the loss values may exhibit a different trend, since, e.g., as the average gets smaller, the variance may instead increase (as in the case of  $F_2$, blue curve).
\end{enumerate}
Thus, upon establishing learning strategies for a given ML task, it is  critical to account for all relevant aspects, including the dependability of the trained models, besides  the training computational cost or the robustness of the compressed model.   
To properly tackle training dependability, in the following we  develop a system model accounting for the stochastic behavior of the test loss evolution, and 
envision a solution strategy that selects the best learning strategy to  achieve a target accuracy \emph{with a target probability}.

\section{System Model and Problem Formulation\label{sec:system-model}}

\textbf{System model.}
We tackle a distributed learning scenario with a set of network nodes, including cloud, edge, and  far-edge nodes, that can contribute to a learning task, and a set of data sources that can transfer data to them. 
All such nodes are assisted by a learning orchestrator located at an edge server, which, given an ML task at hand, selects (i)  the data to be leveraged for the model training, (ii) the ML model to be used, as well as (iii) the set of nodes, hereinafter referred to as learning nodes, that should contribute to the training of such a model. 
To encompass a wide range of distributed learning paradigms, including federated learning (FL) and sequential learning, we consider that the orchestrator identifies clusters of learning nodes, with each cluster performing the vanilla FL~\cite{konen2015federatedOptimization}, and, possibly, contributing to the training process for a given number of epochs before handing the model over to the next cluster. Additionally, each cluster may use a different variation of the ML model (i.e., a full or a compressed version). 
The scenario thus includes the  following main elements: 

\textbullet~Clusters of learning nodes, $n{\in} \Nc$, that, for simplicity and without loss of generality, are assumed to include homogeneous nodes with equal communication and computational capability. We thus denote with  $r(n)$ the computational capability of the generic node in cluster $n$ and with $b(n)$ the bandwidth available on the link  between the FL coordinator within cluster $n$ and each learning node. Each  unit of computing resource has a {\em cost},~$\kappa(n)$, which depends on the equipment available at the cluster nodes;

\textbullet~Datasets, $d{\in}\Dc$, each generated by a data source, that can be conveyed to node clusters. Let $\beta(d,n)$ be the total network cost to transfer dataset $d$ from the source to the nodes\footnote{Again, for simplicity and without loss of generality, we consider that, given $d$, each cluster node receives an equal portion of the dataset.} 
of cluster $n$ before the cluster starts training the model; 

\textbullet~ML models suitable for the task at hand, in their full or compressed versions. They are represented as  elements of the models set, $m{\in}\Mc$. Each model is associated with a size $s(m)$ and a complexity factor.  It follows that, at each epoch, the communication latency within cluster~$n$ due to the learning  nodes sending  their updated model $m$ to their FL coordinator is~$\frac{s(m)}{b(n)}$. Also, given the model complexity, let~$c(k,m,d)$ indicate the quantity of computational resources needed by a learning node to train an epoch $k$ of  model $m$ using data from dataset $d$.

Given a model~$m{\in}\Mc$ and  being cluster $n{\in}\Nc$  the one currently training $m$ with dataset~$d{\subseteq}\Dc$, we 
 define as~$l^\text{run}(k,d,m)$ the global change in the loss  achieved by training model~$m$ over $d$ for the $k$-th epoch. Notice that training aims at minimizing the loss, hence, $l^\text{run}$~will be negative if the training does progress. 
Instead, switching from model~$m(k-1)$ to model~$m(k)$ changes the loss by~$l^\text{sw}(k,d,m(k-1),m(k))$, with~$k$ being the current epoch and~$d\subseteq\Dc$ the dataset being used. Notice that~$l^\text{sw}$ is typically  positive, as model switching often results -- in the short term -- in an increase of the loss.  
Importantly, all these values are {\em input} information to our problem; as discussed in \Sec{fit}, estimating them is an orthogonal problem to our own.

{\bf Decisions and problem formulation.}
The orchestrator has to make the following decisions:

\textbullet~Whether cluster~$n{\in}\Nc$ contributes to the learning at epoch~$k$ or not, expressed through binary variable~$y(k,n){\in}\{0,1\}$;

\textbullet~Whether cluster~$n{\in}\Nc$ uses dataset $d{\in} \Dc$  for learning, expressed through binary variable~$z(d,n){\in}\{0,1\}$;

\textbullet~The amount of resources allocated for the model training by a node of a selected cluster, expressed through the real variable~$x(k,n){\leq} y(k,n)r(n)$;

\textbullet~The total number~$K$ of epochs to run;

\textbullet~For each epoch, the model version $m(k) {\in}\Mc$ to use.

Given the above decisions and
neglecting the one-time data transfer from sources to clusters, 
the duration of epoch~$k$ is  given by the computation and communication time of the selected cluster, i.e.,
\begin{equation}
\label{eq:duration}
T(k)=\frac{c(k,d,m(k))}{x(k,n)}+\frac{s(m(k))}{b(n)}\,.
\end{equation}

Concerning the test loss~$\ell(k)$, it can be computed recursively by accounting for the two $l$-contributions, i.e.,
\begin{equation}
\label{eq:ell-k}
\ell(k){=}\ell(k-1){+}l^\text{run}(k, d,m(k)) 
{+}l^\text{sw}(k,d, m(k{-}1),m(k))\,. \nonumber
\end{equation}
where $\ell(0)$ is the initial loss value. 
Note how the last term in the above equation reduces to zero if no model switching occurs from epoch $k{-}1$ to epoch $k$.

The objective of the orchestrator is to minimize the total cost, subject to the fact that the learning quality target  is achieved within the target time: 
\begin{eqnarray}
\min_{x,y,z,K,m} &  \sum\limits_{k=1}^K \sum\limits_{n{\in}\Nc} \left( \sum\limits_{d{\in} \Dc} \beta(d,n) z(d,n) +  \kappa(n)x(k,n)\right) & \label{eq:obj} \\
\text{s.t.}&   \ell_\omega(K) \leq \ell_\omega^{\max}& \label{eq:constr-ell}\\
& \sum_{k=1}^K T(k)  \leq T^{\max} &\label{eq:constr-t}\\
& x(k,n)\leq y(k,n)r(n) \quad \forall k,n &\label{eq:constr-x}\\
& z(d,n)\leq y(k,n)\quad\forall d, n, k.&\label{eq:constr-z}
\end{eqnarray} 
It is important to highlight that in \Eq{constr-ell} $\omega {\in}[0,1]$ indicates  a specific {\em quantile} of the loss, as opposed to a  loss expected value. By tweaking $\omega$, we are therefore able to enforce different levels of risk, depending upon the scenario and application at hand.
Also, notice how \Eq{constr-z} implies that inactive nodes neither receive nor  use datasets.

As proven in \Prop{hardness} later, directly optimizing the problem above is prohibitively complex, as the problem is NP-hard. Accordingly, to efficiently and effectively solve large-scale problem instances, we propose a heuristic solution  named {\em Dependable Learning (DepL)}, as described below.

\section{The DepL Solution}
\label{sec:depl}
The DepL solution effectively tackles  two critical issues of the system under study. 
First, the need for dependable training, seeking to keep a target quantile~$\omega$ of the loss  distribution below a threshold value~$\ell^{\max}_\omega$ that is deemed to be acceptable for the application at hand. Second, the complexity of the decisions to make, which, along with the sheer amount of existing options  to explore, makes finding the optimal solution impractical for realistic-sized problem instances.  
To address the first issue, DepL  makes  three main decisions:
\begin{enumerate}
    \item selecting the {\em datasets} to use for learning;
    \item choosing the {\em model} to use at each epoch, switching models as needed;
    \item identifying the cluster of {\em learning nodes} to call for at each epoch, and the resources they should commit.
\end{enumerate}
Then, to tackle the second issue, DepL   efficiently makes orchestration decisions by {\em decoupling} them, while  {\em iterating} as needed so as to account for their mutual influence. 

To this end, it is critical to establish the order in which decisions are made. DepL properly accounts for the decisions' mutual influence by making decisions that strongly influence each other sequentially. We observe that the mutual influence is strongest between (i) dataset and model selection, since different models may call for different  data, and (ii) model selection and resource allocation, as insufficient resources may cripple the model performance. 
Additionally, DepL seeks to make first decisions with the deepest impact on the overall training performance. To do so, we remark that:

\textbullet~Dataset selection is related to the model selection {\em and} affects the resources needed at the nodes, as more data invariably means more processing at the nodes;

\textbullet~Model selection is less critical, yet very relevant as it impacts both the learning progress (hence, how many epochs we need) and the required computational resources  (hence, how long each epoch will take);

\textbullet~The impact of the resources and nodes to use is limited to the epoch duration and cost.

\begin{figure}
\centering
\includegraphics[width=1\columnwidth]{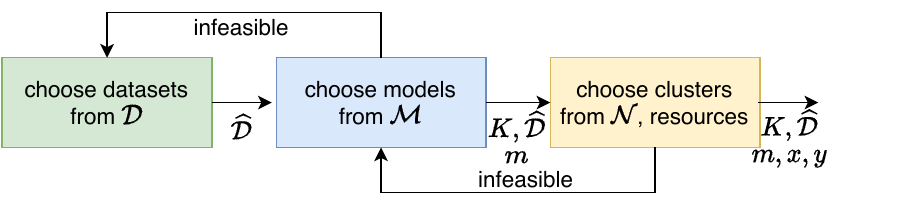}
\caption{
The main steps of the DepL solution strategy.
\label{fig:steps}
}
\vspace{-4mm}
\end{figure}

Based on the above, DepL takes the approach depicted in \Fig{steps}, making dataset selection decisions in the outermost loop, then model selection, and finally resource allocation. 
Importantly, a further reason to make decisions in the order highlighted in \Fig{steps} has to do with how difficult it is to {\em estimate} their consequences. Indeed, choosing the data sources to use for learning  requires sophisticate statistics~\cite{li2021sample} and feedback loops~\cite{wang2020optimizing}. The impact of model selection on the overall learning is somehow better understood, though with significant uncertainty~\cite{noi-infocom23}, and finding the right resources to run a given model can be mapped to many well-studied problems~\cite{sallam2019joint,harris2022dynamic}. By {\em making harder-to-estimate decisions less often, we can reduce the impact of estimation errors} on the overall solution quality.

Furthermore, in view of the above observations, we make each decision by considering the {\em least} restrictive option for the subsequent ones. As an example, while choosing the data, we assume that the most complex (i.e., full) version  of the model is chosen in the subsequent step, and then that the cluster with the largest amount of available resources is used and they   are fully allocated to the model training.  Indeed, since adding more data is the most consequence-fraught decision to make, we want to avoid doing that, unless it is utterly necessary; similarly, a simpler model is used only if the available computational and networking resources are insufficient to make a more complex model work.

\subsection{Dataset selection}
\label{sec:depl-dataset}

Adding new datasets to improve learning quality  
may have undesirable consequences: (i) adding a new dataset implies that  more training should be performed (hence, more resources are needed at the nodes using the dataset);
(ii) more datasets typically imply more network resources to consume  for their collection {\em and}, hence, a larger network. 
Accordingly, we begin with no selected datasets and add them as needed.   
Importantly, at this stage, we take the learning time as proxy of the overall training cost, as we do not directly tackle resource allocation.

Let us denote the set of  datasets  selected for learning with $\widehat{\Dc}=\{d{\in}\Dc\colon\,\exists\, n{\in}\Nc\colon z(d,n){=}1\}$. Let  also  $\widehat{K}(\widehat{\Dc})$ denote the number of epochs needed 
to reach convergence if the datasets  in~$\widehat{\Dc}$ are used, and by~$\widehat{T}(\widehat{\Dc})$ the duration of each epoch. Notice that considering at this stage a fixed resource allocation implies that the epoch duration depends only on the selected datasets.

We  remark that adding a new dataset to~$\widehat{\Dc}$ has two contrasting effects on the training epochs. On the one hand, more learning is performed at each epoch -- hence, fewer epochs are needed to reach the target learning quality. On the other hand, more data is collected and processed at each epoch -- hence, each epoch lasts longer. Our key observation is that these two effects do not have the same evolution:
(i) Adding more data decreases~$\widehat{K}(\widehat{\Dc})$) according to a logarithm law~\cite{linjordet2019impact,sun2017revisiting};
(ii)
Processing and time to transmit model updates, $\widehat{T}(\widehat{\Dc})$ grow instead  linearly with the quantity of data.
%
It follows that, intuitively, if adding a new dataset does not help (specifically, it does not make convergence faster), adding another one will {\em also} not help. More formally, 
$\widehat{K}(\widehat{\Dc})\widehat{T}(\widehat{\Dc})$ is {\em submodular}.

Recalling that here we take the learning time as proxy of the cost, and that the only decision we have to make is selecting the datasets, we write the problem  to solve as:
\begin{equation}
\label{eq:prob-step1}
\min_{\widehat{\Dc}\subseteq\Dc}\widehat{K}(\widehat{\Dc})\widehat{T}(\widehat{\Dc}),
\end{equation}
which is an {\em unbounded submodular minimization problem}~\cite{lovasz1983submodular}. 
 Such problems can be solved to optimality in polynomial time via several different algorithms. Examples range from the traditional ellipsoid method~\cite{lovasz1983submodular} (based on the notion of Lov{\'a}sz extension~\cite{lovasz1983submodular}, linking submodular combinatorial problems to convex continuous ones) to more modern approaches~\cite{itoko2007computational} that achieve {\em strongly polynomial} runtime, only depending upon the problem size.

In summary, regardless of the concrete algorithm used to solve \Eq{prob-step1}, it is possible to choose the datasets in such a way that the learning time (hence, cost) is minimized. This means that DepL solves to optimality the first step of the solution strategy, although not original problem in \Eq{obj}, which is NP-hard (as per \Prop{hardness}). 

\subsection{Model selection}
\label{sec:depl-model}

The high-level goal is now to select a model in~$\Mc$ to use at each epoch. To cope with the complexity and the combinatorial nature of the decision to make, we leverage an {\em expanded graph} approach, predicated upon:
\begin{enumerate}
    \item Creating a graph whose vertices represent the possible states of the system;
    \item Adding edges therein representing the possible decisions and their effects;
    \item Seeking the  path (hence, a sequence of decisions) that connects the current state with a feasible one and  has  the lowest weight (i.e., cost).
\end{enumerate}
As mentioned above, at this stage the datasets to use are given, and we assume that the most capable  node cluster, and all the resources therein, can be used.

To build the expanded graph, let  $\eta$ be an integer parameter determining the accuracy with which we  discretize learning time and test loss. Each vertex of the graph is then associated with a 4-uple~$(m(k),k,T(k),\ell_\omega(k))$, where:
\begin{itemize}
    \item $m(k)$ is a model in~$\Mc$;
    \item $k$ is an epoch in~$1,\dots,K$;
    \item $T(k){=}\frac{i}{\eta}T^{\max}$, with~$i{=}1,\dots,\eta$, is the elapsed learning time in the state represented by the vertex;
    \item $\ell_\omega(k){=}\frac{j}{\eta}\ell(0)$, with $j{=}1,\dots,\eta$ and $\ell(0)$ being the initial loss value,  reflects, in a similar way, the $\omega$-quantile of the current loss.
\end{itemize}

The total number of vertices is then~$|\Mc|K\eta^2$. Such vertices represent {\em a subset} of all possible states of the system (e.g., there is no vertex with~$T(k)=\frac{0.5}{\eta}T^{\max}$), while their indices keep track of the corresponding elapsed training time. This allows the size of the graph to be manageable, at the expense of precision.

Concerning edges, we add an edge from vertex~$(m(k),k,T(k),\ell_\omega(k))$ to vertex~$(m(k+1),k+1,T(k+1),\ell_\omega(k+1))$ if (i) it is possible to switch from~$m(k)$ to~$m(k+1)$, (ii) it takes at most~$T(k+1){-}T(k)$, and (iii) the loss changes by at most~$\ell_\omega(k+1){-}\ell_\omega(k)$. The weight of such edge represents  the cost associated with the switch. Edges are also created between vertices corresponding to the same model, i.e., indicating no model switch.

Finally, we add a virtual vertex~$\Omega$ and a zero-weight edge from any vertex representing a feasible solution (i.e., with an elapsed time lower than~$T^{\max}$ and a loss quantile lower than~$\ell^{\max}_\omega$) to~$\Omega$. Since vertices connected to~$\Omega$ correspond to feasible states, and edges in the graph are a conservative representation of the outcome of  model-switching decisions, any path connecting the current state to~$\Omega$ corresponds to a feasible set of decisions. The lowest-weight among such paths represents then the lowest-cost, i.e., {\em optimal}, decisions we can make, given the graph representation.
Such decisions may not be optimal solutions to the original problem, due to the fact that not all states are represented by vertices in the graph. However, by increasing~$\eta$, we can reduce this issue and get arbitrarily close to the optimum, at the cost if increasing the size of the graph and the solution complexity.

There is an additional problem to deal with, stemming from the fact that expanded-graph approaches work with {\em additive} constraints~\cite{martin2021kpi,xue2007finding}, i.e., constraints where the effect of a sequence of decisions (in our case, model selection at different epochs) can be expressed as the sum of the effects of individual decisions (in our case, the single $l^\text{run}$~components). Such a property is required in order to ensure that the features of a path on the expanded graph (i.e., the associated learning time or loss quantile) can be expressed as the sum of features of individual edges therein. The property holds 
for the elapsed time, and it would be true for the expected loss, but not for the loss quantile. 
Indeed, we know the effects of model selection decisions as distributions, e.g., through their probability density function (pdf); however, the pdf of the sum of two random variables is not the sum of the individual pdfs, but their convolution.

We face this hurdle by adopting the following approach to deal with the evolution of the value of loss-quantile, based on function transforms:
\begin{enumerate}
    \item We move to the transformed domain, replacing pdfs with their Laplace transforms;
    \item Convolutions between pdfs are thus replaced by products of transformed pdfs;
    \item We further compute the logarithm of the transformed pdfs, so that products can be computed as sums.
\end{enumerate} 
It is important to highlight that the aforementioned approach works in both cases (i) when pdfs of the test loss are known exactly (in which case the transforms and logarithms can be computed symbolically), and (ii) when they are not (in which cases all operations are performed numerically). In the latter case, the fast Fourier transform (FFT) may be used {\em in lieu} of the Laplace transform, owing to the vast availability of fast, high-quality implementations.

Finally, if no model can be selected that meets the target learning time and quality, DepL goes back to the previous stage:  the current dataset choice is blacklisted 
and  the datasets selection is repeated,  still considering  the most complex model. Optimized model selection will then be attempted again under the new data selection.

\subsection{Node and resource allocation}
\label{sec:depl-node}

Once the datasets and models are chosen, the only remaining decisions to make concern: (i) the clusters to use for learning, and (ii) which resources each cluster has to commit.
Importantly, neither decision affects how many epochs are needed for convergence, nor the distribution of the resulting loss value; on the other hand, clusters and resource selection have a very strong influence on the learning time (i.e., the duration of each epoch) and the resulting cost.

We make the key observation that the resulting problem -- including its decisions and their effects -- can be mapped, one-to-one, to the classic problem of VNF placement, also referred to as service embedding. Specifically:
(i) clusters  are mapped to  servers, which can run VNFs;
(ii) models  are mapped to the VNF themselves;
(iii) model complexity is mapped to VNF requirements. 
As a consequence, we can use any VNF placement strategy to solve this stage of our own problem. VNF placement is a very well studied problem, with several efficient and effective schemes existing for it. In particular, we  take~\cite{feng2017approximation} as a reference, and obtain the same $O(1+\varepsilon)$~competitive ratio and $O(1/\varepsilon)$ complexity.

Again, in the case where no cluster or resource allocation  can be found that can make the model be trained within the target time, DepL goes back to the previous decision stage, i.e., model switching, again 
blacklisting the model choices that have proven insufficient and 
considering that any cluster and resource can be used. The overall process depicted in \Fig{steps} may therefore be iterated till a feasible solution is found, or the existing options have been exhausted.

\section{Problem and Solution Analysis}
\label{sec:analysis}

In the following, we formally prove several important properties about the problem we solve and our DepL solution. Our results can be summarized as follows:
\begin{itemize}
    \item The problem we solve is NP-hard (\Prop{hardness}), hence, a heuristic solution is required;
    \item The DepL solution has low, namely, polynomial computational complexity (\Prop{efficiency});
    \item The solutions yielded by DepL are provably very close to the optimum (\Prop{effectiveness}).
\end{itemize}

Beginning from the problem, we prove its NP-hardness via a reduction from the knapsack problem.
\begin{property}
\label{prop:hardness}
The problem of optimizing \Eq{obj} subject to constraints \Eq{constr-ell}--\Eq{constr-z} is NP-hard.
\end{property}
\begin{IEEEproof}
We prove NP-hardness by reduction from the knapsack problem, which is known~\cite{kellerer2004introduction} to be NP-hard. To do so, we provide a polynomial-time procedure transforming any instance of the knapsack problem into an instance of our problem.
Recall that the input to the knapsack problem consists of a set~$\Ic{=}\{i\}$ of items, each with weight~$w_i$ and value~$v_i$, along with a maximum weight~$w^{\max}$. We have to decide whether to take each item~$i$ with the goal of maximizing the total value, subject to the constraint that the total weight does not exceed~$w^{\max}$. 
Given the above, we can create an instance of our problem where:
    (1) items correspond to models in~$\Mc$;
    (2) the learning improvement yielded by each model is deterministic and equal to the value of the corresponding item;
    (3) the requirements of each model are equal to the cost of the corresponding item;
    (4) there is only one dataset and only one cluster, with sufficient capabilities to run any model;
    (5) all data transfer costs~$\beta(d,n)$ are set to zero.
It can be seen by inspection that reduction is polynomial in complexity -- indeed, constant, as it has no loops. As we have successfully reduced an instance of an NP-hard problem to an instance of our own, we can conclude that our problem is NP-hard.
\end{IEEEproof}
It is also worth remarking that the instance of our problem generated by the reduction is a very simple one; this suggests that our problem is significantly more complex than the knapsack problem in practice. It also implies that general-purpose heuristic approaches for the knapsack problem (or any other known problem) will work poorly with ours, rather, domain- and problem-specific strategies like DepL are warranted.

Concerning DepL, we can prove that it has a remarkably low, namely, quadratic worst-case computational complexity:
\begin{property}
\label{prop:efficiency}
The DepL procedure has worst-case polynomial complexity of~$O\left(|\Dc|^2+|\Mc|^2K^2\frac{1}{\eta^4}+\frac{1}{\varepsilon}\right)$.
\end{property}
\begin{IEEEproof}
The total complexity is obtained by considering the three steps depicted in \Fig{steps} and described in \Sec{depl}.

\begin{figure*}[h] 
    \centering
    \includegraphics[width=0.32\textwidth]{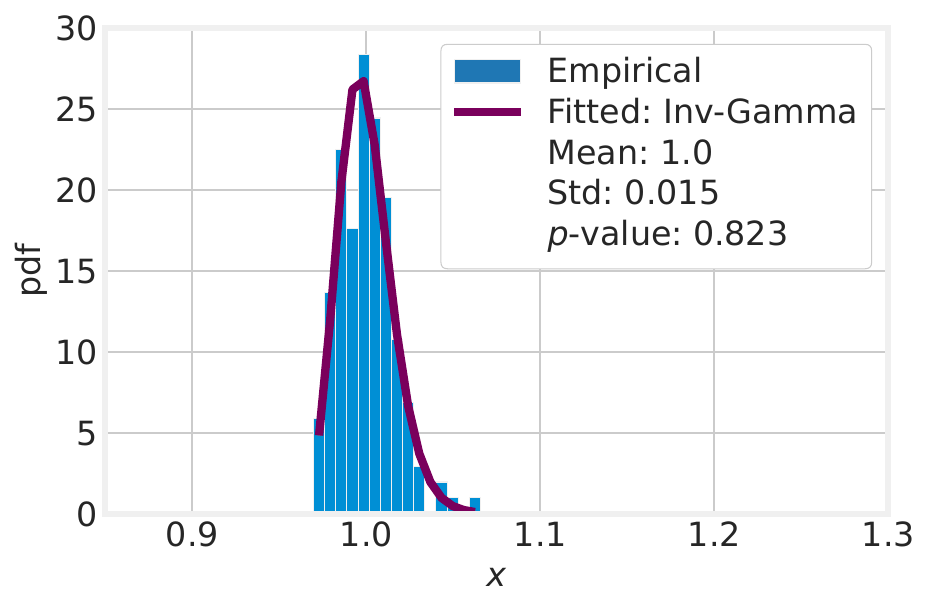}
    \includegraphics[width=0.32\textwidth]{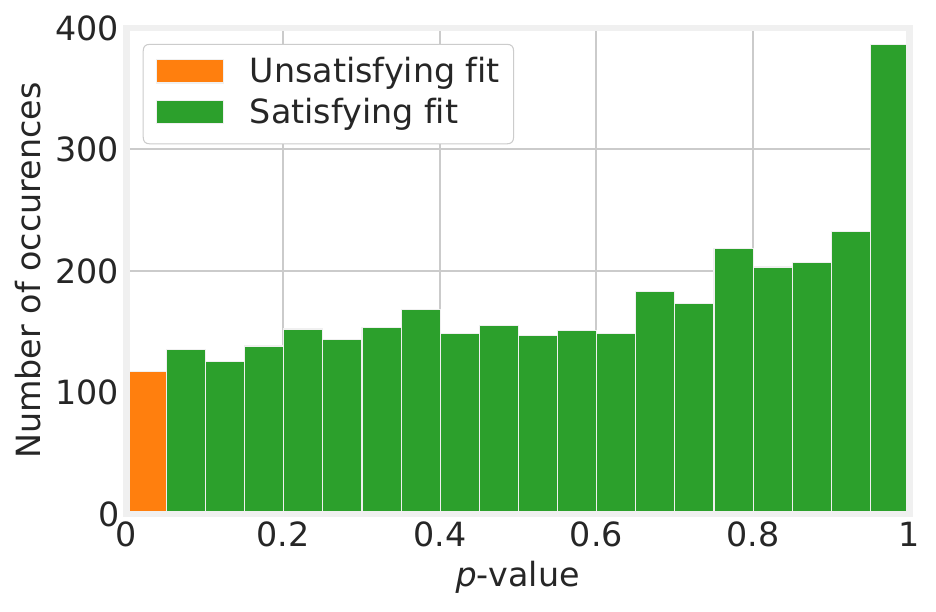}
    \includegraphics[width=0.32\textwidth]{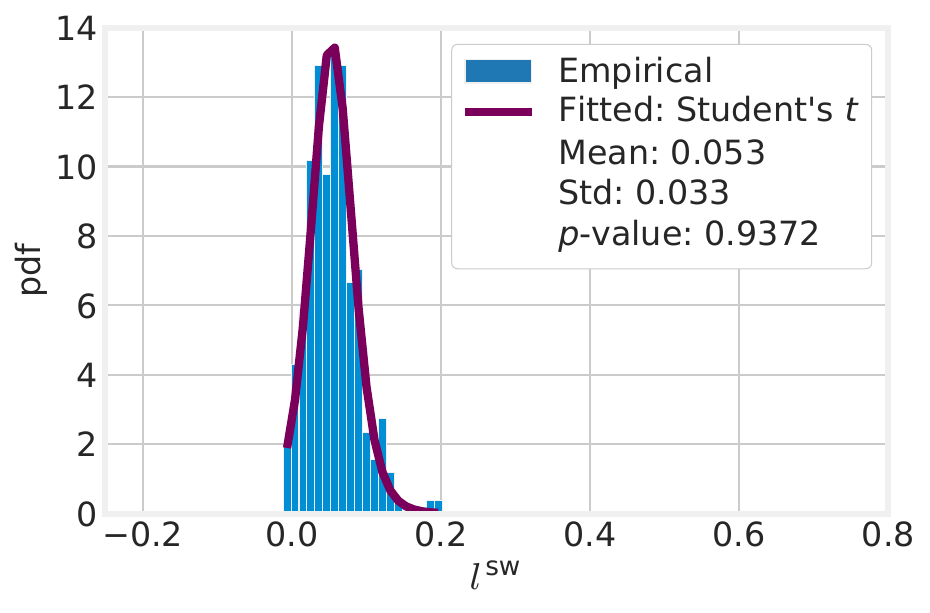}
    \caption{
    Example distributions of~$\Xc$ as defined in \Eq{model} along with the Inverse-Gamma fit (left); distribution of the $p$-values highlighting fitting quality (center); example distribution of~$l^\text{sw}$ as defined in \Eq{ell-k} along with the Student's t fit (right).
    \label{fig:distributions}
} 
\end{figure*}

For dataset selection (\Sec{depl-dataset}), we leverage the results in~\cite{itoko2007computational}, solving unbounded submodular minimization problems to the optimum with a computational complexity that is quadratic in the size of the set over which to optimize, i.e.,~$\Dc$. 
For model switching (\Sec{depl-model}), we have to compute the minimum-cost path over the expanded graph. The maximum number~$V$ of vertices in the expanded graph is given by all combinations of model, epoch, time, and loss, hence,
$V\leq |\Mc|K\frac{T^{\max}}{\eta}\frac{\ell^{\max}}{\eta}$. 
Recalling that minimum-cost path algorithms, e.g., Dijkstra's, have a complexity of~$O(V^2)$, it follows that the complexity of the model selection procedure is~$O\left(|\Mc|^2K^2\frac{1}{\eta^4}\right)$. 
Finally, for cluster selection (\Sec{depl-node}), we rely on~\cite{feng2017approximation}, which has a complexity of~$O\left(\frac{1}{\varepsilon}\right)$, where $\varepsilon$~is a configurable optimality parameter. 
Combining all contributions, we obtain
$O\left(|\Dc|^2+|\Mc|^2K^2\frac{1}{\eta^4}+\frac{1}{\varepsilon}\right)$,
which proves the thesis.
\end{IEEEproof}

The last aspect we look at is the effectiveness of the decisions made by DepL. We can prove that the competitive ratio (i.e., the ratio of DepL's solution cost to the optimal one) is constant (i.e., it does not depend upon the problem size) and remarkably small.
\begin{property}
\label{prop:effectiveness}
The DepL solution has a  constant competitive ratio, namely,~$\frac{1}{\eta}(1+\varepsilon)$.
\end{property}
\begin{IEEEproof}
Similarly to the proof of \Prop{efficiency}, we need to consider the competitive ratio of each of the steps described in \Sec{depl}, and then multiply them. 
Dataset selection (\Sec{depl-dataset}) is optimal, hence, the corresponding competitive ratio is~$1$. For model selection, instead, the only source of sub-optimality is the parameter~$\eta$; specifically, as per~\cite[Theorem~4.3]{xue2007finding}, the resulting competitive ratio is~$\frac{1}{\eta}$. Finally, the approach in~\cite{feng2017approximation}, which we consider as a reference for node selection, yields a competitive ratio of~$1+\varepsilon$.
\end{IEEEproof}
A constant competitive ratio like the one proven in \Prop{effectiveness} implies that the quality of the solutions found by DepL is not influenced by the size of the problem instance being solved; intuitively, DepL is {\em robust} to large problem sizes.

\section{Numerical Results\label{sec:results}}

In this section, we first (\Sec{fit}) demonstrate how the~$l^\text{run}$ and~$l^\text{sw}$ parameters can be estimated through real-world experiments. Then (\Sec{peva}), we leverage the resulting data to compare the performance of DepL, the optimum, and the state-of-the-art.

\begin{figure*}
\centering
\includegraphics[width=.32\textwidth]{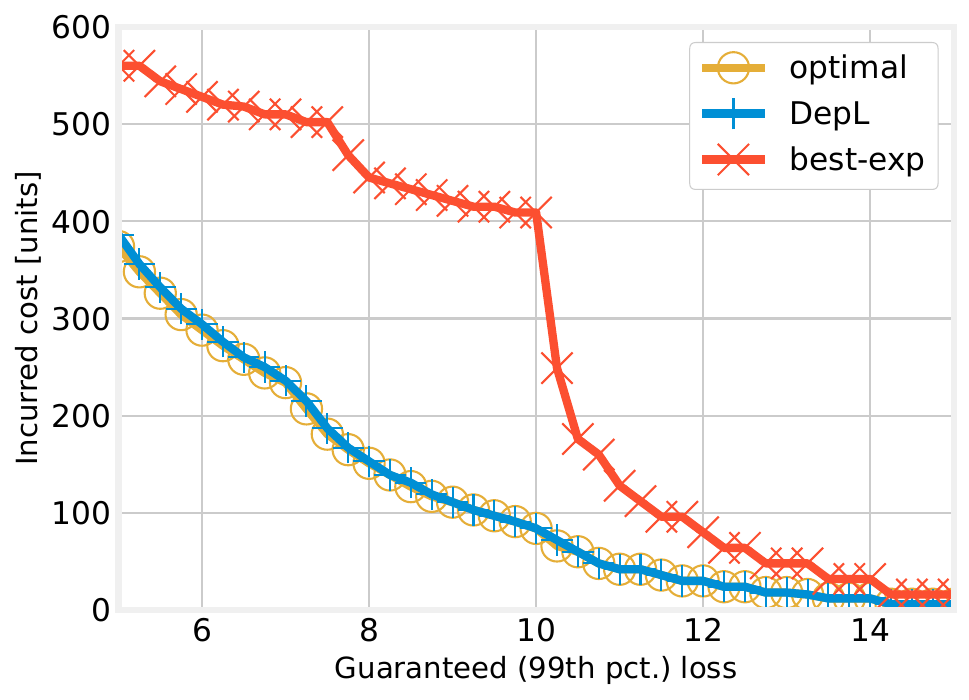}
\includegraphics[width=.32\textwidth]{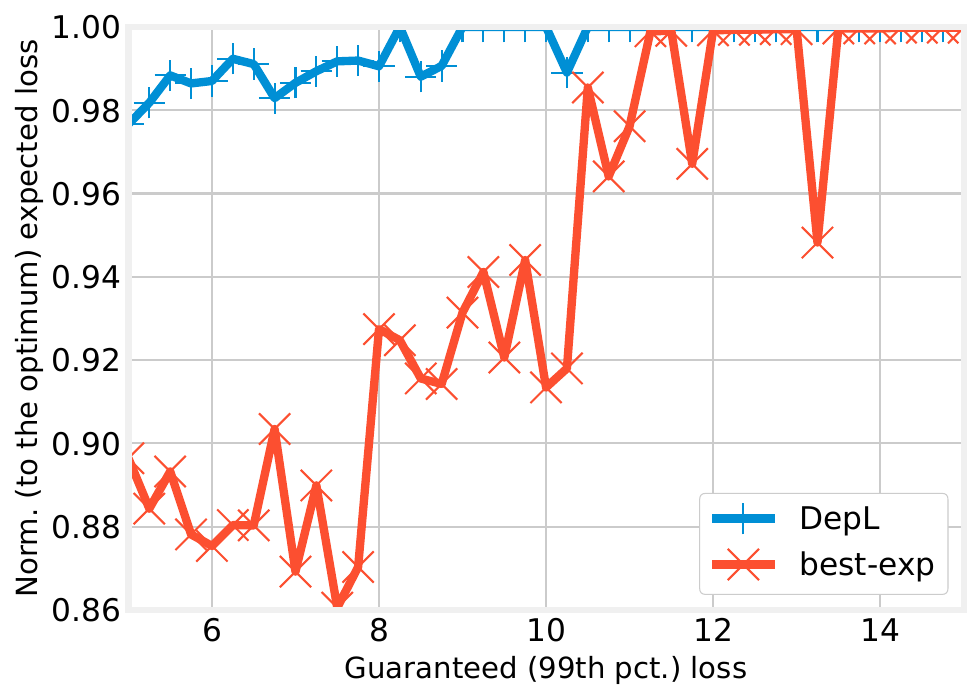}
\includegraphics[width=.32\textwidth]{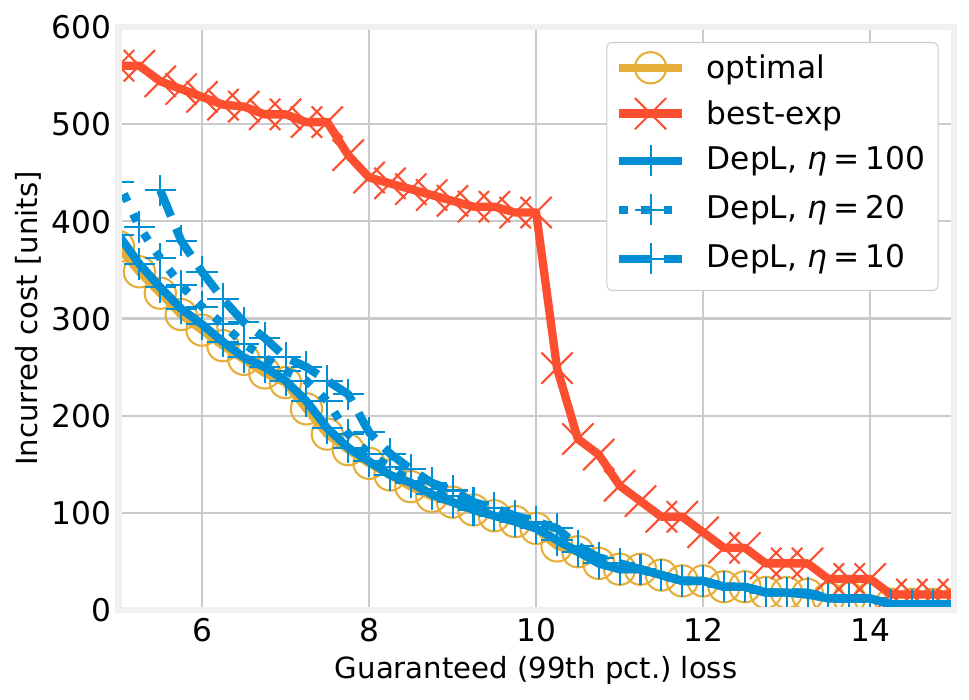}
\caption{AlexNet: performance of DepL and alternative benchmarks: cost (left); normalized expected loss (center); effect of $\eta$ (right).
    \label{fig:peva-basic}
} 
\centering
\includegraphics[width=.32\textwidth]{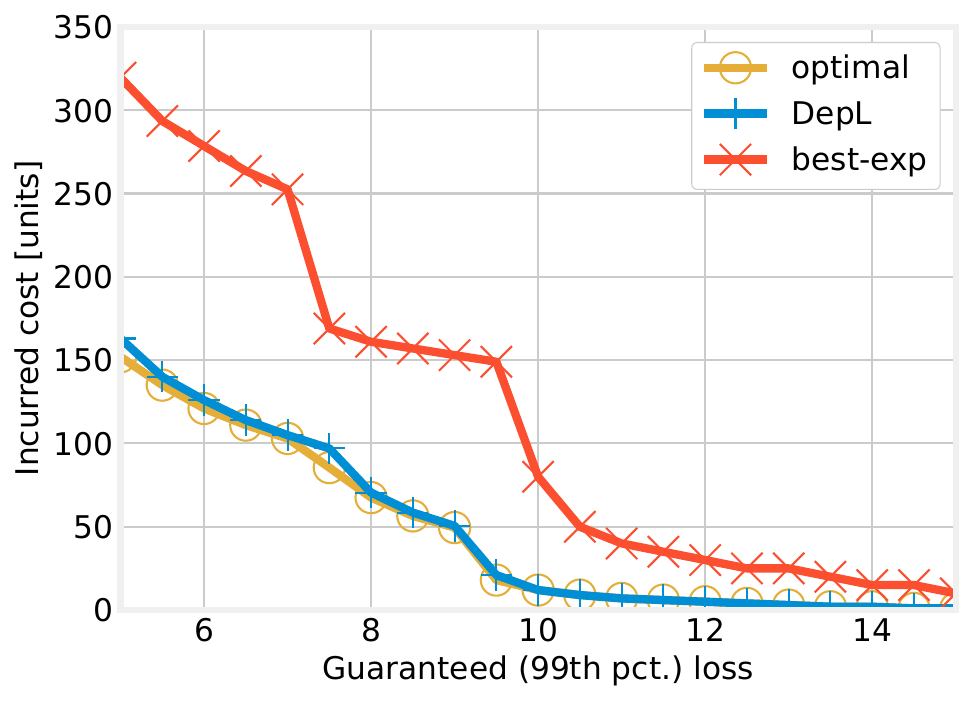}
\includegraphics[width=.32\textwidth]{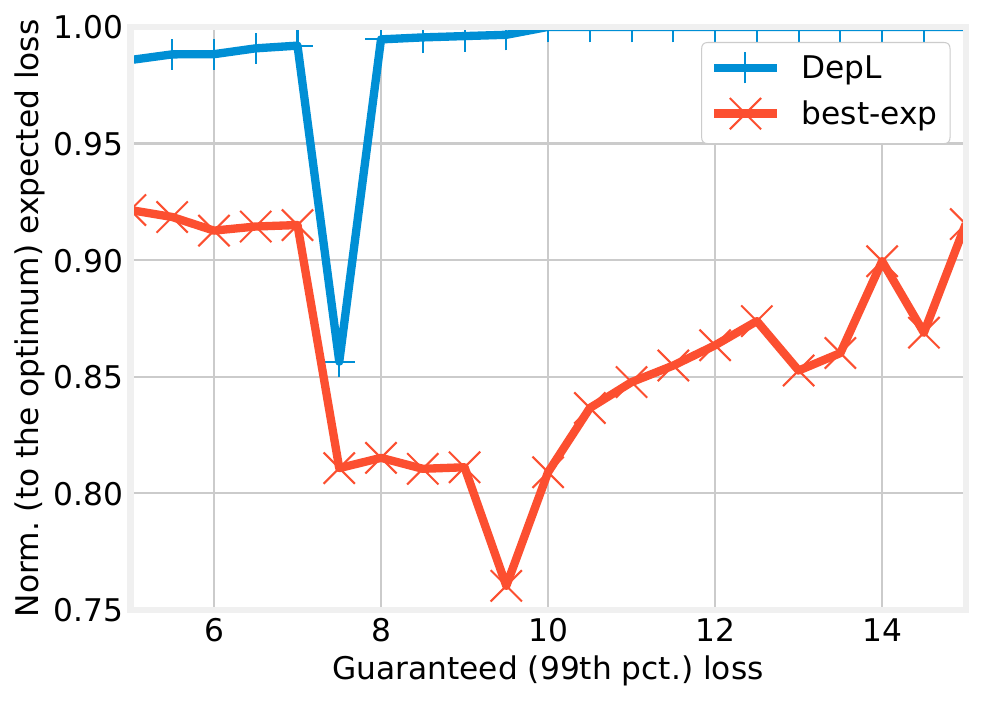}
\includegraphics[width=.32\textwidth]{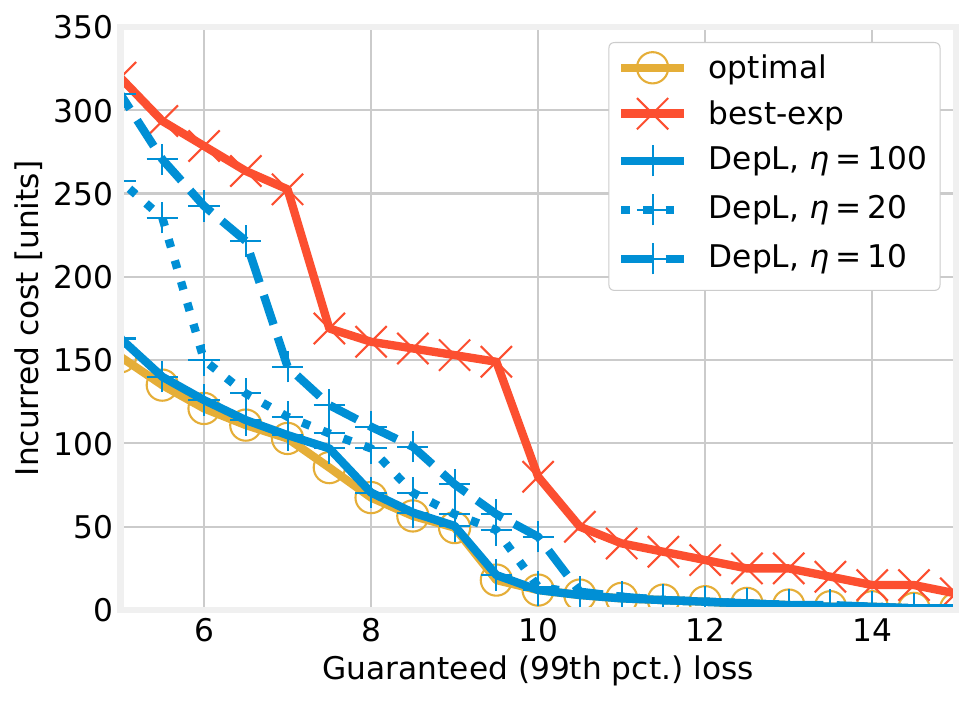}
\caption{
MobileNet: performance of DepL and alternative benchmarks. Cost (left); normalized expected loss (center); effect of $\eta$ (right).
    \label{fig:peva-basic2}
} 
\end{figure*}
\begin{figure*}
\centering
\includegraphics[width=.32\textwidth]{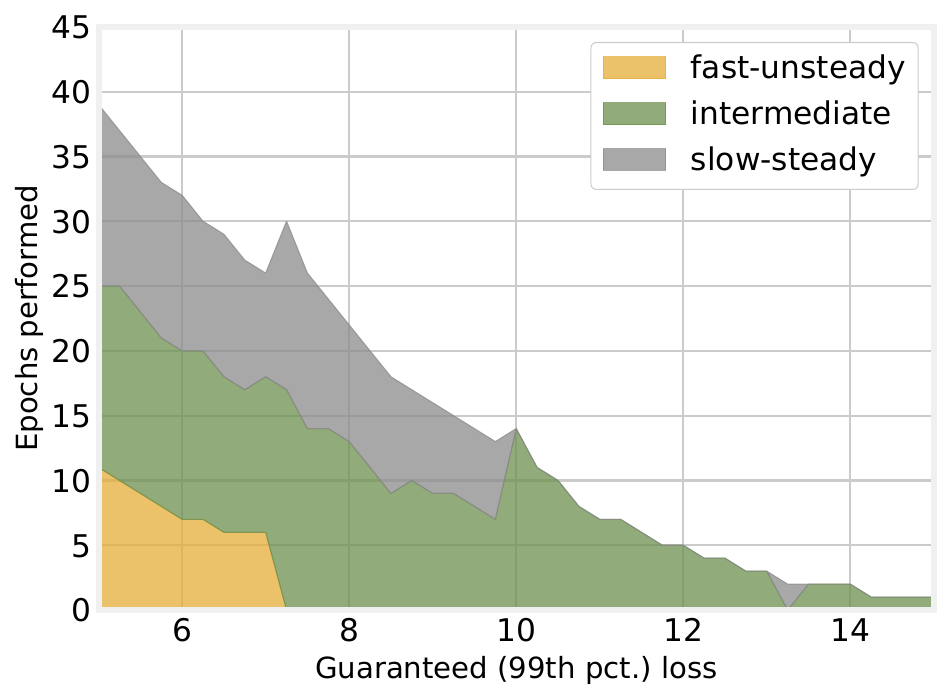}
\includegraphics[width=.32\textwidth]{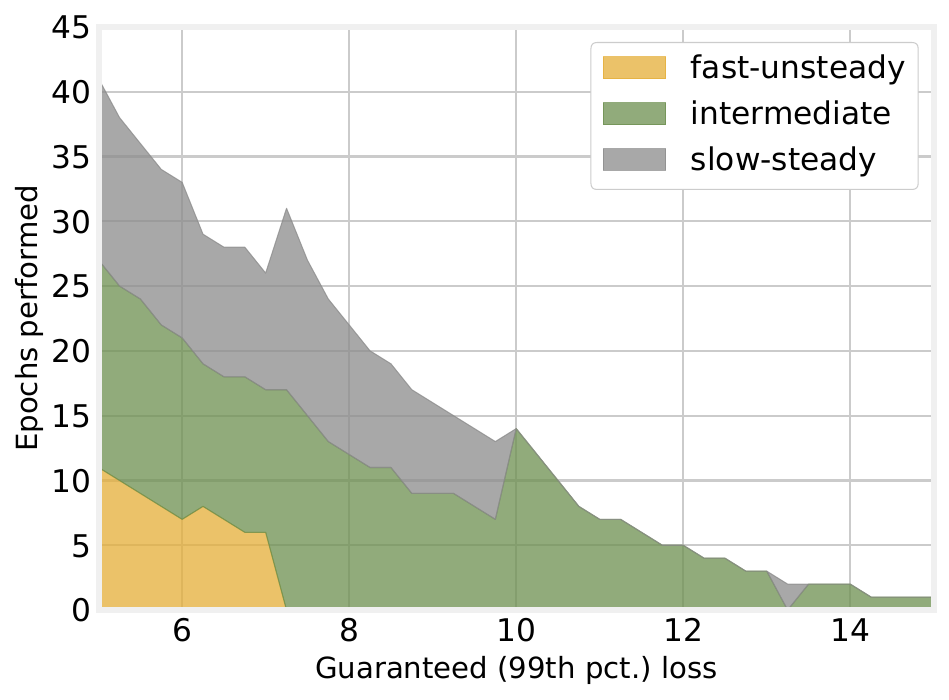}
\includegraphics[width=.32\textwidth]{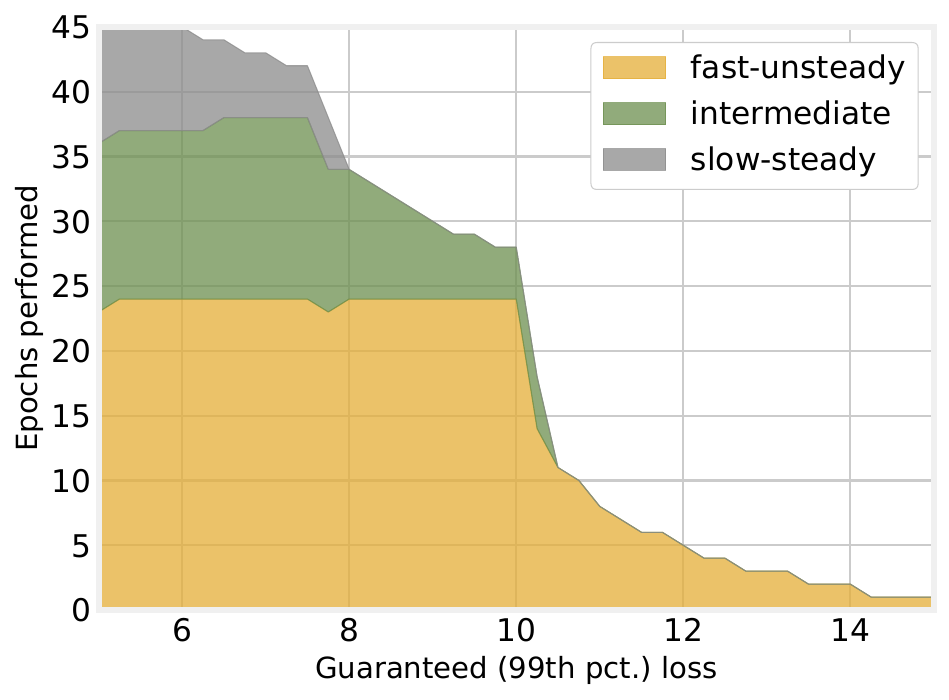}
\caption{Epochs spent with each model by the optimal solution (left), DepL (center), and {\em best-exp} (right).
    \label{fig:peva-epochs}
} 
\vspace*{-4mm}
\end{figure*}

\subsection{Estimating the loss evolution}
\label{sec:fit}

As observed in \Sec{motivation}, the evolution of the test loss
over the training epochs, $k$, cannot be predicted with certainty, rather,  a probabilistic representation thereof should be used. Previous work~\cite{li2019convergence} has shown that the loss evolution can be well approximated by an inverse-square-root law, i.e., 
it is proportional to~$\frac{1}{\sqrt{k}}$. 
We generalize such an expression 
as:
\begin{equation}
\label{eq:model}
\ell^\text{run}(m,k)=\frac{A_m}{\sqrt{k+B_m}}\cdot x,
\end{equation}
where~$x\sim\Xc_m$ is a sample drawn from distribution~$\Xc_m$, 
which is specific to the considered model and data. 
The parameters in \Eq{model} can be estimated as follows:
  (i) $A_m$ and~$B_m$ are obtained through traditional parameter fitting techniques, e.g., least-squares, under the assumption that~$x{=}1$;
  (ii) given~$A_m$ and~$B_m$, $\Xc_m$ is obtained through distribution fitting.

\Fig{distributions}(left) shows the empirical pdf of $x$ for an example test loss curve, along with the fitted pdf $\Xc_m$, in the case where
two clusters of nodes sequentially  train the AlexNet CNN
 using the CIFAR-10 dataset.  
The two clusters are located, respectively, in the network edge and the far-edge, and perform FedAvg. The order in which the clusters contribute to the training goes from the highest (edge) to the lowest (far-edge) computing capability, as ultimately the model needs to be personalized with the data owned by IoT or user devices (which have indeed lower compute capabilities). At each training stage,  the model is further pruned, so as to make it amenable for training at the lower-end nodes.  Let  $K_1$ ($K_2$) be the number of epochs performed at the first (second) stage, using a model pruned with factor $F_1$ ($F_2$); for \Fig{distributions}(left) we set $K_1{=}17$, $K_2{=}60$, $F_1{=}0.5$, and $F_2{=}0.75$. 
The empirical pdf, and, more specifically, the  standard deviation, highlight that, given a combination of learning parameters, the prediction of the next loss value exhibits significant uncertainty, thus further confirming the need for a probabilistic representation of the loss trajectory.  
Importantly, one can notice that the Inverse-Gamma distribution provides an excellent goodness of fit. The latter is measured by computing the \textit{p}-value using the  Kolmogorov-Smirnov (KS) test, with \textit{p}-values higher than   $0.05$ corresponding to a good fit\footnote{A \textit{p}-value higher than  $0.05$ indicates that our null hypothesis, i.e., data are distributed according to an Inverse-Gamma distribution, cannot be rejected.}. 
The histogram in \Fig{distributions}(center) shows the \textit{p}-values for the distributions obtained  using all possible combinations of the number of epochs  $K_1 \in [1,20]$ and $K_2 {\in} \{1,5,15,25,40,60\}$, and of the pruning factors $F_1,F_2 {\in}\{0.25, 0.36, 0.5, 0.75\}$: in almost all cases they are larger than the $0.05$ threshold, again indicating the goodness of fit of the Inverse-Gamma distribution on the empirical distributions.

Model switching, e.g., compression, has a  different effect: namely, a one-time increase in the loss (hence, a reduction in accuracy). We modeled this effect  by $l^\text{sw}$, i.e., the difference between the test loss values after and before the model switching. Clearly, each combination of learning parameters yields a different distribution. 
\Fig{distributions}(right) shows an example of $l^{sw}$ distribution, representing the empirical and the fitted pdfs for $F_1{=}0.36$, $F_2{=}0.5$.
The empirical pdf, in \Fig{distributions}(right), shows that the support of $l^{sw}$ distribution is composed of  positive values, indicating an increase of the test loss after model switching. Again, such an increase cannot be predicted with certainty; hence, we need a probabilistic representation of $l^{sw}$. 
In this case, the KS test reveals that  satisfying \textit{p}-values can be obtained using the Student's $t$-distribution.

Below, we use the above experimental estimation of the loss evolution to assess and benchmark the performance of DepL.

\subsection{Performance evaluation}
\label{sec:peva}

Since, as discussed in \Sec{depl-dataset}, the dataset  selection
is provably optimal, 
our performance evaluation focuses on model and node selection, i.e., steps~2 and 3 in \Fig{steps}. To this end, we benchmark our DepL solution described in \Sec{depl-model} against two alternatives:
\begin{itemize}
    \item Optimal decisions, obtained by brute force  (labelled as {\em optimal} in plots);
    \item A state-of-the-art solution, aiming to optimize the expected loss, inspired to~\cite{noi-infocom23}  (labelled as {\em best-exp} in plots).
\end{itemize}
The  best-exp scheme makes  decisions of the same nature as our DepL; however, it accounts for the expected loss instead of any quantile thereof. Thus, comparing against best-exp is a valuable way to gauge the impact of accounting for the distribution of the loss besides its expected value. 
Finally, unless otherwise specified, for DepL we set~$\eta{=}200$.

For all strategies, we consider as items of~$\Mc$:
\begin{itemize}
    \item either three versions of the AlexNet DNN pruned, respectively, with factors~$F_1{=}0.5$, $F_2{=}0.75$, and $F_3{=}0.9$,
    \item or three versions of the MobileNet network, pruned according to the same factors.
\end{itemize}
\Tab{models} summarizes the features of the DNNs we use.
Specifically, as the name assigned to the models suggests, the ``fast-unsteady'' 
model provides a fast learning on average, but it is liable to occasionally yield poor test loss performance;  
``slow-steady'', on the other hand, provides slower but more reliable convergence, while ``intermediate'' yields  balanced performance.

We consider a network infrastructure including two types of server (hence, clusters nodes in~$\Nc$):
 A-class nodes, with capabilities set to~$r(n){=}100$ and processing cost~$\kappa(n){=}1$, and B-class nodes, with~$r(n){=}50$ and processing cost~$\kappa(n){=}1.2$. For simplicity, we assume that clusters contain~one node each, and set~$\beta(n,d)=0$ for all nodes and sources.
For all strategies, node assignment and resource allocation decisions are made following the methodology in~\cite{feng2017approximation}.

The first aspect in which we are interested  is the relative performance of the strategies we compare, i.e., how good they are at minimizing the objective \Eq{obj}. The results
for the AlexNet model
are summarized in \Fig{peva-basic}(left), showing the cost incurred for different target values of the loss 99th percentile. As one might expect, tighter loss requirements require more training, hence, a higher cost. Most interestingly, DepL  outperforms best-exp and virtually matches the optimum. The difference is especially large when the requirements are {\em less} strict, hence, there is more margin for learning strategies that optimize cost over sheer performance.

\begin{table}[t]
\caption{
AlexNet and MobileNet versions used for performance evaluation: expected and variance of loss improvement per epoch, 
and resource demand\label{tab:models}
\vspace{-2mm}} 
\begin{tabularx}{\columnwidth}{|l||X|X|X||X|X|X|}
\hline
Model & Exp. loss improv. & Var. of loss imp. & Norm. needed resources & Exp. loss improv & Var. of loss imp. & Norm. needed resources \\ \hline
& \multicolumn{3}{c||}{AlexNet} & \multicolumn{3}{c|}{MobileNet}\\ \hline
Fast-unsteady & $1.02$ & $0.031$ & 2.6 & $1.02$ & $0.029$ & 1.8\\ \hline
Intermediate & $1.00$ & $0.010$ & 1.4 & $0.95$ & $0.012$ & 1.0\\ \hline
Slow-steady & $0.98$ & $0.007$ & 1 & $0.92$ & $0.006$ & 0.6\\ \hline
\end{tabularx}
\vspace{-4mm}
\end{table}

\begin{figure*}
\centering
\includegraphics[width=.32\textwidth]{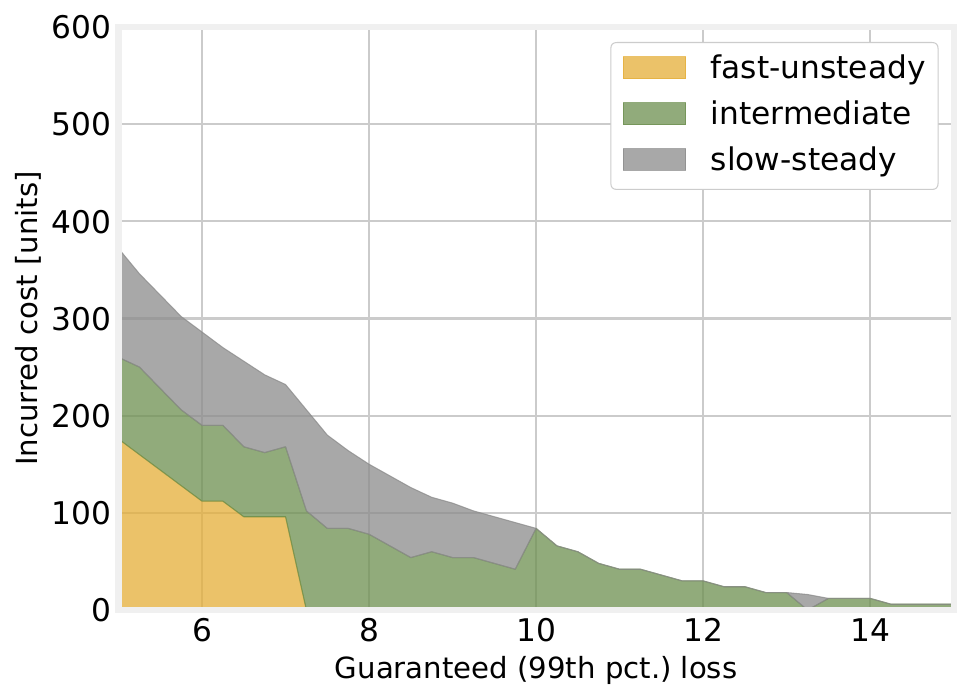}
\includegraphics[width=.32\textwidth]{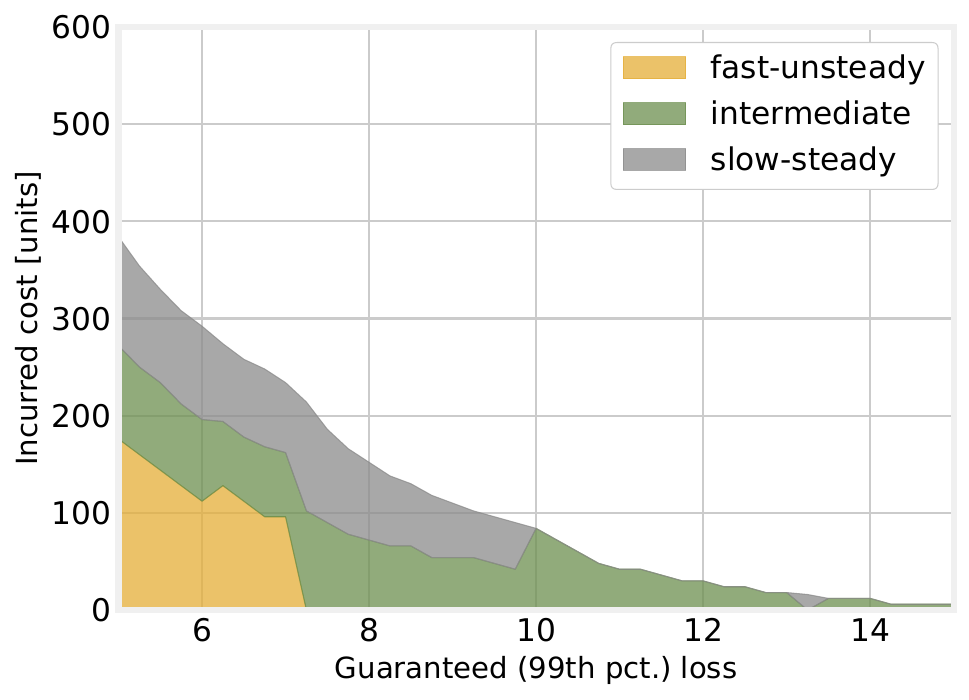}
\includegraphics[width=.32\textwidth]{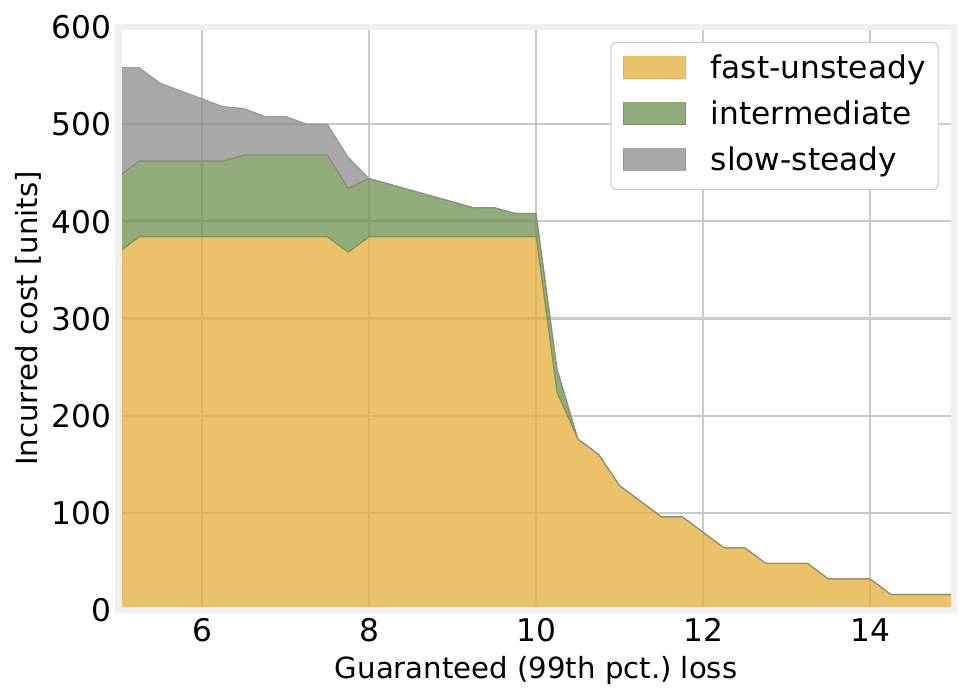}
\caption{
    AlexNet: cost incurred with each model by the optimal solution (left), DepL (center), and {\em best-exp} (right).
    \label{fig:peva-resources}
} 
\centering
\includegraphics[width=.32\textwidth]{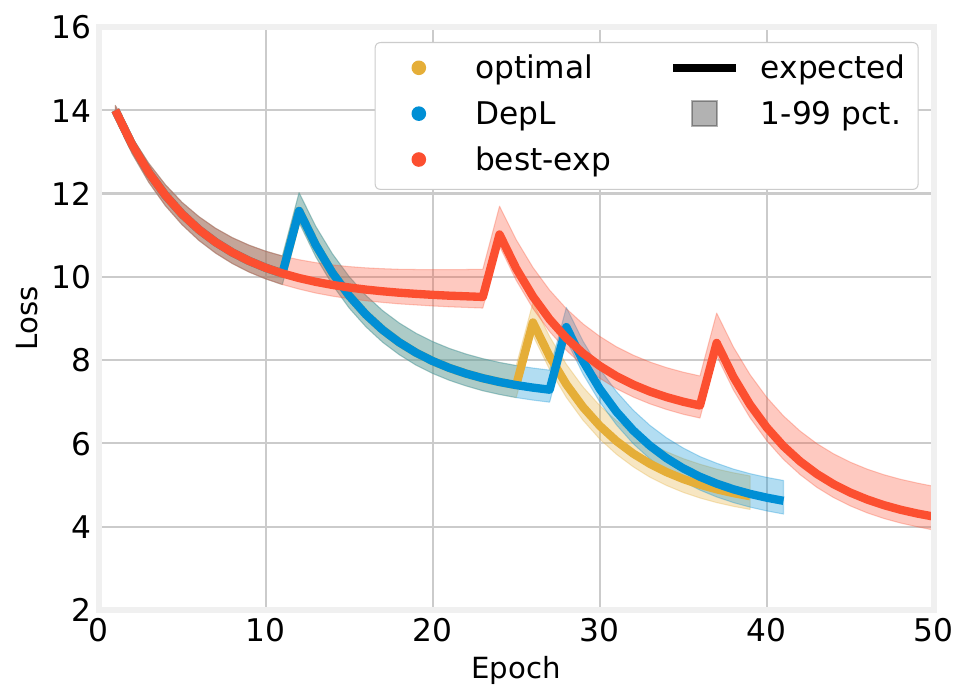}
\includegraphics[width=.32\textwidth]{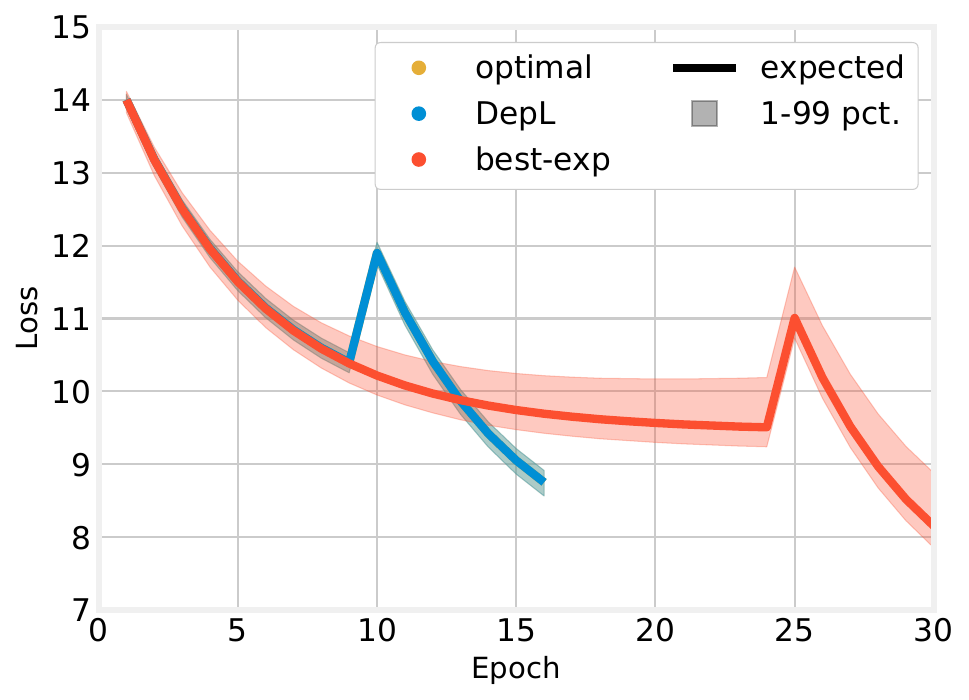}
\includegraphics[width=.32\textwidth]{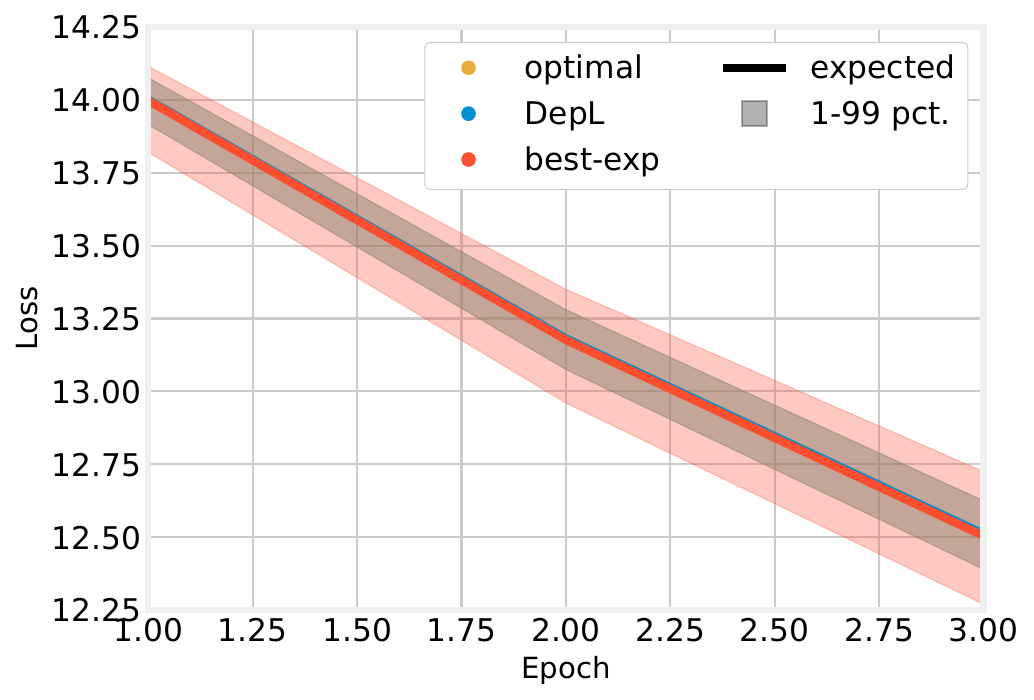}
\caption{
    AlexNet: evolution of the expected and guaranteed loss when the guaranteed loss is $5$~(left), $9$~(center), and $13$~(right).
    \label{fig:peva-trajectories}
} 
\vspace{-4mm}
\end{figure*}

\Fig{peva-basic}(center), depicting the loss achieved by DepL and  best-exp,
normalized to the optimum, 
sheds additional light on this effect.  All strategies meet the learning quantile constraint \Eq{constr-ell} with an equality sign, i.e., they provide exactly the required value of the loss 99th percentile. On the other hand, \Fig{peva-basic}(center) shows that both  best-exp and, to a much smaller extent, DepL yield a lower (i.e., better) expected loss than the optimum. Clearly, this is a waste of resources: since our constraint is on the 99th percentile of the loss, improving its expected value brings no benefit.

In \Fig{peva-basic}(right), we assess the effect of decreasing the parameter~$\eta$, which we use when building the expanded graph in \Sec{depl-model}, on the performance of DepL. We observe
how there is indeed a significant margin to pursue different trade-offs between DepL's computational efficiency and the quality of the decisions it yields, as indicated by the analysis in \Sec{analysis}.

The same behavior can be observed from \Fig{peva-basic2}, presenting the behavior when MobileNet versions are used. We can observe that, compared to \Fig{peva-basic}, costs tend to be lower, a testament to the better efficiency of MobileNet compared to AlexNet. Furthermore, costs decrease more steeply as the loss target increases, showing that MobileNet struggles more for more complex learning tasks. Besides these differences, it is important to remark how -- exactly like in \Fig{peva-basic} -- DepL significantly outperforms best-exp, and closely matches the optimum.

Moving back to AlexNet, in
\Fig{peva-epochs} we take a closer look at the decisions made by each strategy, specifically, at the number of epochs for which each of the models in \Tab{models} is used for training. DepL (center plot)  makes very similar choices to the optimum (left plot), consistently with \Fig{peva-basic}; in both cases, all models are used as required by the loss quantile  target. Interestingly, when the target is loose, both strategies can only use the ``intermediate'' model, and avoid expensive model switches. On the contrary, as the target gets tighter, switches become necessary and all models are used. The right plot, depicting the decisions made by  best-exp, shows a completely different picture. As best-exp optimizes the expected loss, it utilizes the ``fast-unsteady'' model as much as possible, 
which results in a higher number of epochs to run, hence, 
the higher costs presented in \Fig{peva-basic}(left).

\Fig{peva-resources} shows the total cost of each strategy, broken down according to the model they use. As in \Fig{peva-basic}(left),  best-exp  
requires the most computational resources, hence, incurs the highest 
cost; also, in accordance with \Tab{models}, much of such cost is due to the ``fast-unsteady'' model. This confirms how solely focusing on the expected loss means performing more epochs (\Fig{peva-epochs}) {\em and} spending more in each of them (\Fig{peva-resources}), while obtaining a needlessly low expected loss (\Fig{peva-basic}(center)).

Finally, \Fig{peva-trajectories} shows the evolution of the loss across different epochs, for different strategies and values of the target learning quality; peaks correspond to model switches. In all plots, lines represent the expected loss, while the shaded area is delimited by the loss 1st and 99th percentiles. We can immediately observe that the lines corresponding to DepL (and the optimum) are shorter, i.e., fewer epochs are necessary to reach the target learning quality. Further, it is evident (especially in the middle plot) how  best-exp is associated with wider shaded areas, i.e., a larger distance between the loss 1st and 99th percentiles. This is consistent with \Fig{peva-epochs} and \Fig{peva-resources}: from those figures, we can observe that best-exp tends to use more the ``fast-unsteady'' model, which is associated with a larger variance (as per \Tab{models}). Furthermore, \Fig{peva-trajectories} confirms how DepL virtually always matches the optimum, making model switching decisions that account for both the expected value {\em and the distribution} of the loss.

\section{Related work}
\label{sec:relwork}

DepL  is related to three main research areas: resource-aware distributed DNN training, DNN model compression, and robust DNN training.
In the former case, the goal is to locate and exploit the best resources for a given training task. The main decision is often  choosing the nodes to involve in training, based on  their speed~\cite{wang2019adaptive,abdelmoniem2021resource}, the quantity~\cite{marfoq2022personalized,abdelmoniem2021resource} and quality~\cite{wu2021fast} of their data, the speed and reliability of their network~\cite{wang2019adaptive,zhou2021communication}, as well as trust~\cite{imteaj2020fedar}. In all cases, the model to train is assumed to be given and fixed. 
Recent works~\cite{malandrino2022energy} target more complex scenarios, where layers of the DNN can be duplicated at different learning nodes if needed.  The studies in~\cite{paissan2022scalable,noi-infocom23} also seek to adapt the learning task to the available resources, choosing from different existing models. 

Model compression is a set of techniques to transfer the knowledge from a model to a different (usually simpler) one. Prominent among such techniques is model pruning, which consists in removing some weights from a DNN model,  
assuming that very few of the parameters of a model actually impact its performance. Choosing the right parameters to prune is a challenging problem itself; many solutions adopt an  iterative approach~\cite{iterative-pruning-Tan}, by observing the evolution of the parameter values. Other compression techniques include knowledge distillation~\cite{gou2021knowledge}, where a ``student'' model is trained to mimic the decisions of a ``teacher'' model. 
Later studies have tackled how distillation can be combined with generative models~\cite{huang2018distilling} and domain adaptation~\cite{gou2021knowledge}.

Robustness in distributed DNN training is an emerging topic. Most works set in a FL scenario and focus on node selection.
Such a robustness objective is balanced against fairness consideration in~\cite{li2021ditto}, aiming at distributing the training load as evenly as possible. Robustness considerations are also combined with communication efficiency in~\cite{ang2020robust} and data quality.
Importantly, all the aforementioned works deal with node selection; the reliability of the DNN model and the training process are never accounted for.

An alternative approach to adding reliability to DNN results is the so-called {\em conformal prediction}~\cite{angelopoulos2022conformal,zecchin2023forking}. Conformal approaches create a set of possible results of a given DNN (e.g., predicted class) such that the correct result (e.g., the true class) lies in the set with a target probability; intuitively, better models will have smaller conformal sets. Conformal prediction techniques are applied {\em after} training, hence, they can be seamlessly combined with DepL, and indeed benefit from the high-quality training it yields.


\section{Conclusions\label{sec:concl}}
We presented, analyzed and evaluated DepL, a strategy for the dependable training of distributed ML models. Unlike previous work, DepL can guarantee that the target learning quality (e.g., accuracy) is reached {\em with a target probability}. DepL achieves this goal  by making joint, high-quality decisions about the data, nodes, and models to use, including the time at which model switching (through, e.g., compression) shall be performed.
We have formally proved that DepL achieves near-optimal decisions with a constant competitive ratio and low  complexity. We have further shown that  DepL closely matches the optimum and significantly outperforms the state-of-the-art.

\section*{Acknowledgment}
This work was supported by the SNS-JU-2022 project ADROIT6G under the European Union’s Horizon Europe research and innovation programme under Grand Agreement No.\,101095363.
\bibliographystyle{IEEEtran}
\bibliography{refs}

\end{document}